\documentclass{article} % For LaTeX2e
\usepackage{iclr2024_conference,times}

\usepackage{amsmath,amsfonts,bm}

\def\eqref#1{equation~\ref{#1}}
\def\1{\bm{1}}

\def\appnabla{{\widehat{\nabla}}}

\def\vzero{{\bm{0}}}

\def\vtheta{{\bm{\theta}}}
\def\vomega{{\bm{\omega}}}
\def\vpi{{\bm{\pi}}}

\def\vx{{\bm{x}}}
\def\vy{{\bm{y}}}

\def\mD{{\bm{D}}}

\def\mG{{\bm{G}}}

\def\mI{{\bm{I}}}

\def\mS{{\bm{S}}}

\def\mZero{{\bm{0}}}
\def\mOne{{\bm{1}}}

\DeclareMathAlphabet{\mathsfit}{\encodingdefault}{\sfdefault}{m}{sl}
\SetMathAlphabet{\mathsfit}{bold}{\encodingdefault}{\sfdefault}{bx}{n}

\DeclareMathOperator*{\argmax}{arg\,max}

\usepackage{hyperref}
\usepackage{multirow}
\usepackage{enumitem}
\usepackage{graphicx}
\usepackage{booktabs}  
\usepackage{url}
\usepackage{enumitem}
\usepackage{amsmath}
\usepackage{caption}
\usepackage{subcaption}
\usepackage{wrapfig}
\usepackage{tabularx, booktabs}
\newcolumntype{Y}{>{\centering\arraybackslash}X}
\newcolumntype{L}{>{\arraybackslash}X}

\usepackage[linesnumbered,ruled,vlined]{algorithm2e}

\title{Sparse Backpropagation for MoE Training}

\author{%
  Liyuan Liu\textsuperscript{$\mathsection$} \;\; Jianfeng Gao\textsuperscript{$\mathsection$} \;\; Weizhu Chen\textsuperscript{$\ddagger$} \\
  {$\mathsection$}Microsoft Research\;\; {$\ddagger$}Microsoft Azure AI\\
  {\small \texttt{\{lucliu, jfgao, wzchen\}@microsoft.com} }\\
}

\newcommand{\ours}{SparseMixer }
\newcommand{\oursst}{SparseMixer-1st }
\newcommand{\oursrd}{SparseMixer-2rd }

\newcommand{\smallsection}[1]{\textbf{#1.~~~~}}

\newcommand{\noise}[1]{implicit label noise\xspace}
\newcommand{\Noise}[1]{Implicit label noise\xspace}

\iclrfinalcopy % Uncomment for camera-ready version, but NOT for submission.
\begin{document}

\maketitle
\begin{abstract}

One defining characteristic of Mixture-of-Expert (MoE) models is their capacity for conducting sparse computation via expert routing, leading to remarkable scalability.
However, backpropagation, the cornerstone of deep learning, requires dense computation, thereby posting challenges in MoE gradient computations.
Here, we introduce SparseMixer, a scalable gradient estimator that bridges the gap between backpropagation and sparse expert routing. 
Unlike typical MoE training which strategically neglects certain gradient terms for the sake of sparse computation and scalability, SparseMixer provides scalable gradient approximations for these terms, enabling reliable gradient estimation in MoE training. 
Grounded in a numerical ODE framework, SparseMixer harnesses the mid-point method, a second-order ODE solver, to deliver precise gradient approximations with negligible computational overhead. 
Applying \ours to Switch Transformer on both pre-training and machine translation tasks, \ours showcases considerable performance gain, accelerating training convergence up to 2 times\footnote{Implementations are available at \url{https://github.com/microsoft/SparseMixer/}.}. 
\end{abstract}

\section{Introduction}

The significant success of large-scale pre-training across various applications has underscored the imperative need for scalable models that are economically feasible~\citep{chowdhery2022palm,openai2023gpt4,touvron2023llama}. 
Recent advances in sparsely activated networks, prominently known as Mixture-of-Experts (MoE), have attracted widespread interest~\citep{Shazeer2017OutrageouslyLN,Lepikhin2020GShardSG,Fedus2021SwitchTS,riquelme2021scaling,mustafa2022multimodal}. 
Unlike traditional networks that densely activate all modules for all input, MoE 
% strategically allocates different modules to specific inputs 
selectively activates parts of modules to specific inputs 
through a process called {expert routing}, leading to notable efficiency enhancements.

% As ultra large scale pre-training achieves remarkable success in various applications~\citep{chowdhery2022palm,openai2023gpt4,touvron2023llama}, it becomes a pressing issue how to scale the model to the next stage in an affordable manner. 
% Recent advances in sparsely activated networks, also known as Mixture-of-Experts (MoE), have attracted extensive attention~\citep{riquelme2021scaling,mustafa2022multimodal,Shazeer2017OutrageouslyLN,Lepikhin2020GShardSG,Fedus2021SwitchTS}. 
% Instead of using all modules for all inputs, MoE assigns different parts of modules for different inputs (referred to as \emph{routing}), yielding a significant boost in efficiency. 

% Besides great potential for increasing training efficiency, MoE models bring significant challenges to gradient estimation for the sparse modules. 
However, such efficiency gain comes at a cost: gradient estimation in MoE becomes challenging due to expert routing. 
Specifically, the routing function, being discrete in nature, produces non-differentiable outputs. 
Meanwhile, backpropagation, the cornerstone of deep learning, relies on the Chain rule, making it exclusively compatible with differentiable functions~\citep{rosenblatt1957perceptron,Bengio2013EstimatingOP}, and cannot be directly applied for gradient computation of expert routing.

% One crucial step towards constructing MoE networks is the gradient estimation of the sparse modules. 
% The routing function yields discrete module outputs, thus not a differentiable function. 
% However, back-propagation, the cornerstone of deep learning, builds upon the Chain's rule and can only be applied to differentiable functions. 

% In fact, the gap between back-propagation and discrete variables is a common issue that is met by lots of applications.
Numerous methods have emerged to bridge discrete and back-propagation, and most of them are based on Straight-Through (ST)~\citep{rosenblatt1957perceptron,Bengio2013EstimatingOP,Jang2016CategoricalRW,Liu2023BridgingDA}. 
Unfortunately, all existing ST estimators are incompatible with MoE, since they require activating all experts for gradient computing, thereby eliminating all the efficiency improvements of MoE. 
% Correspondingly, typical MoE training practice treats the routing results as constants and neglects its gradient computation. 
Consequently, typical MoE training strategically neglects the gradient computation for routing, trading certain training signals for sparse computation. 
% and thus scalability. 
Despite the scalability brought by sparse computation, this trade-off may result in slow convergence and improperly trained models.

% \begin{figure}[b]
%     \centering 
%     \vspace{-2.5mm}
%     \begin{subfigure}[t]{0.194\linewidth}
%         \centering
%         \includegraphics[width=1.0\textwidth]{figure/wmt_e2.pdf}
%         \vspace{-4mm}
%         % \caption{}
%         \label{}
%     \end{subfigure}
%     \begin{subfigure}[t]{0.194\linewidth}
%         \centering
%         \includegraphics[width=1.0\textwidth]{figure/wmt_e4.pdf}
%         \vspace{-4mm}
%         % \caption{}
%         \label{}
%     \end{subfigure}
%     \begin{subfigure}[t]{0.194\linewidth}
%         \centering
%         \includegraphics[width=1.0\textwidth]{figure/wmt_e6.pdf} % to be updated
%         \vspace{-4mm}
%         % \caption{}
%         \label{}
%     \end{subfigure}
%     \begin{subfigure}[t]{0.194\linewidth}
%         \centering
%         \includegraphics[width=1.0\textwidth]{figure/wmt_e8.pdf}
%         \vspace{-4mm}
%         % \caption{}
%         \label{}
%     \end{subfigure}
%     \begin{subfigure}[t]{0.194\linewidth}
%         \centering
%         \includegraphics[width=1.0\textwidth]{figure/wmt_e16.pdf} % to be updated
%         \vspace{-4mm}
%         % \caption{}
%         \label{}
%     \end{subfigure}
%     \vspace{-2.5mm}
%     \caption{
%     Training curves on WMT'14 En-De.
%     }
%     \vspace{-4mm}
%     \label{fig:poly}
% \end{figure}

\begin{figure}[bh!]
    \centering 
    % \vspace{-.5mm}
    \includegraphics[width=1.0\textwidth]{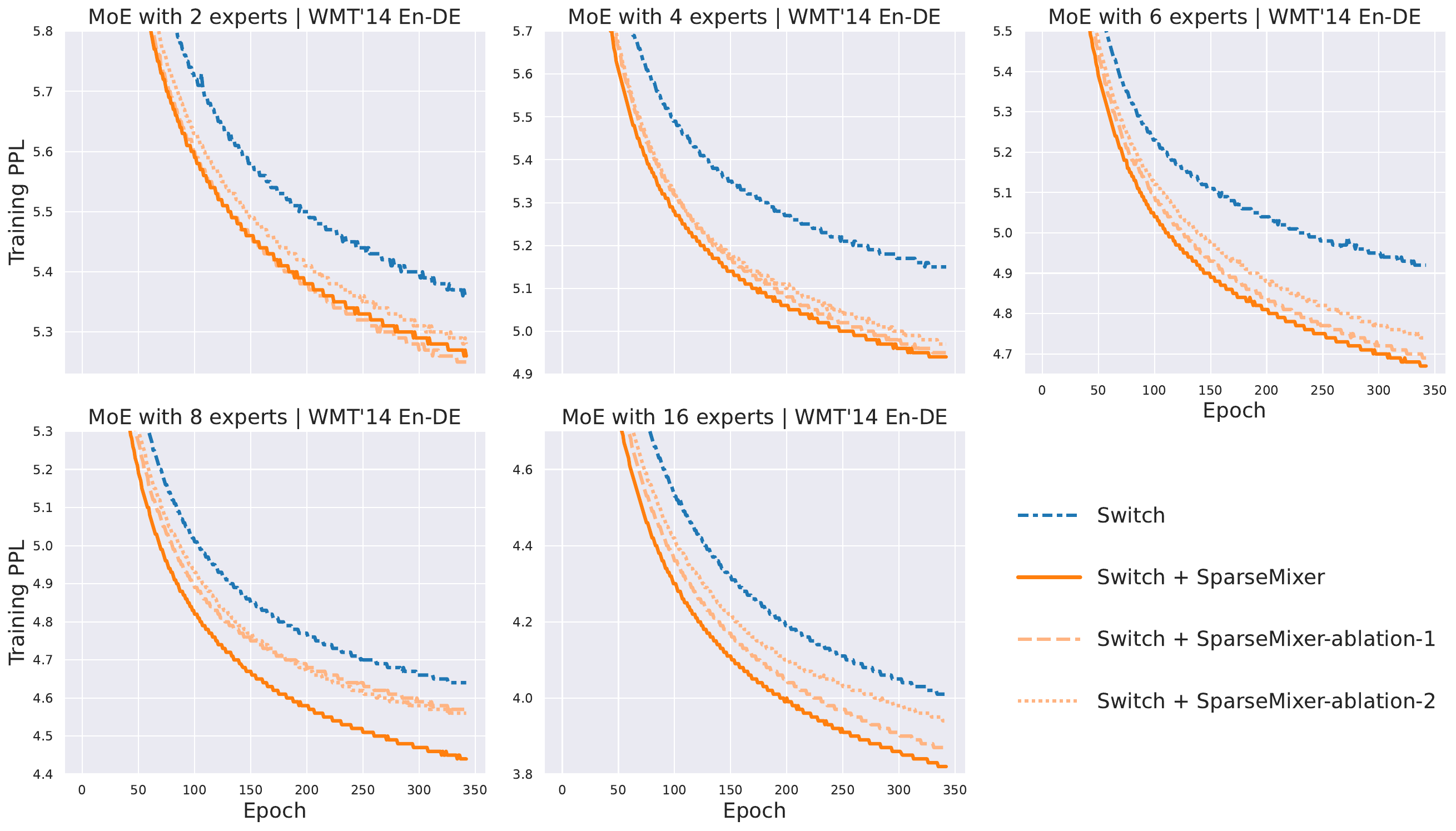}
    \vspace{-6.0mm}
    \caption{
    Training curves of Switch Transformer on WMT'14 En-De.
    }
    \vspace{-4mm}
    \label{fig:wmt14}
\end{figure}

% In this study, we present \ours to move beyond discrete and bridge the gap between sparse MoE routing and back-propagation. 
Our solution to this quandary is SparseMixer—a novel approach designed to reconcile the divide between sparse MoE routing and backpropagation. 
% Distinguishing itself from traditional ST estimators, 
% % Compared to existing ST estimators, 
% \ours offers a novel gradient approximation approach, yielding a first-order gradient with only activated expert outputs.  
% % adopts a different approach to approximate the gradient, thus producing a first-order gradient approximation with the output of only one expert. 
% Adopting a novel numerical ODE perspective, 
% Adopting a novel perspective inspired by numerical methods for ordinary differential equations (ODE),
% \ours makes it possible to compute gradient approximation for expert routing, with only sparsely activated expert outputs. 
Drawing inspiration from numerical methods for ordinary differential equations (ODE), SparseMixer provides reliable gradient approximation for expert routing, even when only a subset of experts are activated. 
Moreover, to furnish accurate gradient approximations with negligible computation overheads, 
we integrate the mid-point method, a second-order numerical ODE solver, which matches the Taylor expansion of the gradient to the second order without requiring the Hessian matrix or other second-order derivatives.

% We apply \ours to Switch Transformer and conduct empirical studies on both pretraining and neural machine translation tasks.
We apply SparseMixer to Switch Transformer on both pretraining and neural machine translation.
\ours not only accelerates training convergence by up to two times but also facilitates MoE with properly trained expert routing. 
Remarkably, while Switch Transformer underperforms the dense model in all three pretraining settings, incorporating \ours as the gradient estimator allows the resulting MoE models to consistently outperform the dense model. 

\section{Related Work and Preliminary}
\label{sect:notation}

\smallsection{Mixture-of-Expert for Transformer}
The idea of Mixture-of-Expert models 
% dates back to 
originates from 
\citet{Jacobs1991AdaptiveMO} and \citet{Jordan1994HierarchicalMO}, which integrates many separate networks and uses each to handle a separate subset of training cases. 
Recently, many attempts have been made to leverage this idea for scaling large language models~\citep{Shazeer2017OutrageouslyLN, Lepikhin2020GShardSG,Lewis2021BASELS,Fedus2021SwitchTS}.

To keep things straightforward, we will first focus on a simplified setting of the switch Transformer layer~\citep{Fedus2021SwitchTS}. 
We will then discuss its difference with the Switch Transformer and necessary adaptations in Section~\ref{subsec:setting}, while the resulting algorithm can be easily extended to other MoE designs. 
Considering a set of N experts, $\{ f_i(\vx) \}_{i=1}^N$, the gate value of expert $i$ is computed with the softmax function as $\vpi_i = \mbox{\small softmax}(\vtheta)_i = \frac{\exp(\vtheta_i)}{\sum_{j=1}^n \exp(\vtheta_j)}$, where $\vtheta = W_r \cdot \vx$. 
For $i \in [1, \cdots, N]$, we mark its one-hot representation as $\mI_i \in \mathcal{R}^{N \times 1}$, whose element equals $1$ if it is the $i$-th element or equals $0$ otherwise. 
Let $\mD$ be a discrete random variable and $\mD \in \{\mI_1, \cdots, \mI_N\}$.
Note that $\mD$ is sampled as $\mD \sim \vpi$ during training, and is computed as $\mD \gets \argmax_{\mI_i} \vpi_{\mI_i}$ during inference. 

Then, the final output of this MoE layer is $\vy = \vpi_\mD f_\mD (\vx)$.
Marking other parts of the neural network as a differentiable function $g: \mathcal{R}^{n} \to \mathcal{R}$, we aim to minimize: % \lucas{add a second sampling process}
\begin{equation}
    \min_{W_r} \mathcal{L}(W_r), \; \mbox{ where } \mathcal{L}(W_r)  = E_{\mD \sim \mbox{\small softmax}(W_r \vx)} [g(\vpi_\mD f_\mD(\vx))] = \sum_\mD \vpi_\mD \cdot g(\vpi_\mD f_\mD(\vx)).
    \label{eqn:objective_as_expectation}
\end{equation}
% We believe our modeling here reflects the mechanism of the switch Transformer, and more discussions are elaborated in Section 
% It is worth mentioning that our modeling of the switch Transformer layer (i.e., Equation~\ref{eqn:objective_as_expectation}) is expressed differently than in many literature.
% We will discuss their difference and necessary adaptations in Section~\ref{subsec:setting}.

\smallsection{Gradient Computation for Expert Routing}
For simplicity, we mark $\frac{\partial \mathcal{L}(W_r)}{\partial W_r}$ as $\nabla_0 + \nabla_1$:
\begin{eqnarray}
    \frac{\partial \mathcal{L}}{\partial W_r} := \nabla_0 + \nabla_1,   
    \mbox{where }\nabla_0 = \sum_{\mI_i} g(\vpi_{\mI_i} f_{\mI_i}(\vx)) \frac{\partial\,\vpi_{\mI_i}}{\partial\,W_r} \mbox{ and } \nabla_1 = \sum_{\mI_i} \vpi_{\mI_i} \frac{\partial g(\vpi_{\mI_i} f_{\mI_i}(\vx))}{\partial\,W_r}. \label{eqn:exact}
\end{eqnarray}
It is easy to notice that $\nabla_1$ can be computed reliably via backpropagation. 
$\nabla_0$, however, 
% cannot be computed via back-propagating through $g$. 
is hard to reliably estimate in typical MoE training practice. 
In this study, we focus our discussions on $\nabla_0$. 

% It is usually too costly to compute $\nabla$, since it requires the computation of $g(\pi_i f_i(\vx))$ and $\frac{\partial g(\pi_i f_i(\vx))}{\partial\,W_r}$ for all $i \in \{1, \cdots, N\}$.  
% Correspondingly, many efforts have been made to estimate $\nabla$ efficiently. 

REINFORCE~\citep{Williams1992SimpleSG} is unbiased (i.e., $E[\nabla_{\mbox{\scriptsize REINFORCE}}]=\nabla_0$) and only requires the distribution of the discrete variable to be differentiable (i.e., no backpropagation through $g$): 
\begin{equation}
    \nabla_{\mbox{\scriptsize REINFORCE}} := g(\vpi_\mD f_\mD(\vx)) \frac{\partial\,\log \vpi_\mD}{\partial\,W_r}.
    \label{eqn:reinforce}
\end{equation}
Despite the  $\nabla_{\mbox{\scriptsize REINFORCE}}$ estimator being unbiased, it tends to have prohibitively high variance, especially for networks that have other sources of randomness (i.e., dropout or other independent random variables). 
Recently, attempts have been made to reduce the variance of REINFORCE~\citep{Gu2015MuPropUB,Tucker2017REBARLU,Grathwohl2017BackpropagationTT,shi2022gradient}.
Still, it has been found that the REINFORCE-style estimators fail to work well in MoE training~\citep{Kool2021UnbiasedGE}. 

\smallsection{Straight-Through}
Despite $\nabla_{\mbox{\scriptsize REINFORCE}}$ being unbiased, it treats the remaining network ($g$) as a black-box and only leverages the zero-order information of $g$. 
In practice, a popular family of estimators, Straight-Through (ST), leveraging the first-order information of $g$ (note that $g$ is a scalar and $g'$ is a vector), has been shown to achieve a better performance in more complicated settings~\citep{Liu2023BridgingDA}. 
ST computes the backpropagation ``through'' a surrogate that treats the non-differentiable function (e.g., the sampling of $\mD$) as an identity function~\citep{rosenblatt1957perceptron,Bengio2013EstimatingOP,Jang2016CategoricalRW,Liu2023BridgingDA}.
In our MoE setting, ST treats the sampling of $\mD$ as an identity function and estimates the gradient as:
\begin{equation}
    \appnabla_{\mbox{\scriptsize ST}} := \frac{\partial g(\vpi_\mD f_\mD(\vx))}{\partial\, \vpi_\mD f_\mD(\vx)} \frac{\partial \sum_{i} \mD_i \vpi_{\mI_i} f_{\mI_i} (\vx) }{\partial \mD}\frac{\partial \vpi_\mD}{\partial W_r}.
    \label{eqn:st}
\end{equation}

An alternative strategy is to conduct the concrete random variable relaxation~\citep{Maddison2014AS, Jang2016CategoricalRW}.  
It is observed that the sampling of $\mD$ can be reparameterized using Gumbel random variables at the zero-temperature limit of the tempered softmax~\citep{Gumbel1954StatisticalTO}: 
\begin{equation}
    \mD = \lim_{\tau \to 0} \mS_\tau, \; \mbox{where } \mS_\tau=\mbox{softmax}_\tau (\vtheta + \mG), \mG_i \mbox{ are i.i.d., and } \mG_i \sim \mbox{Gumbel}(0, 1).
\nonumber
\end{equation}
Straight-Through Gumbel-Softmax (STGS) treats the zero-temperature limit as identity function during the backpropagation:
\begin{equation}
\appnabla_{\mbox{\scriptsize STGS}} := \frac{\partial g(\vpi_\mD f_\mD(\vx))}{\partial\, \vpi_\mD f_\mD(\vx)} \frac{\partial \sum_{i} \mS_{\tau, i} \vpi_{\mI_i} f_{\mI_i} (\vx) }{\partial \mS_{\tau}}\frac{\partial \mS_\tau}{\partial W_r}.
    \label{eqn:stgs}
\end{equation}

% Our recent work formally demonstrates that $E[\appnabla_{\mbox{\scriptsize ST}}]$ is a first-order approximation to $\nabla_0$ and further furnishes gradient estimation, achieving second-order accuracy with negligible computation overheads~\citet{Liu2023BridgingDA}. 

Although $E[\appnabla_{\mbox{\scriptsize ST}}]$ has been formally established as a first-order approximation of $\nabla_0$~\citep{Liu2023BridgingDA}, applying ST estimators necessitates the need for computing $f_i(\vx)$ for all $i \in \{\mI_1, \cdots, \mI_N\}$, i.e., the outputs from all experts.
For example, in Equation~\ref{eqn:st}, we have $\frac{\partial \sum_{i} \mD_i \vpi_{\mI_i} f_{\mI_i} (\vx) }{\partial \mD} = \mbox{diag}(\sum_{i} \mD_i \vpi_{\mI_i} f_{\mI_i} (\vx))$, which involves the computation of $\{f_{\mI_1}(\vx), \cdots, f_{\mI_N}(\vx)\}$.
Essentially, computing all $f_{\mI_i}$ turns MoE into a densely activated network. 
Thus, using ST-style estimators undermines the sparse computation, fundamentally obstructing the scaling of MoE models. 

\smallsection{MoE Training Practice}
Due to all these challenges, the current MoE training practice trades certain training signals for scalability. 
Specifically, $\nabla_0$ is strategically neglected in gradient computation (the value of $\nabla_0$ is set to 0), and only $\nabla_1$ is used for model training~\citep{Fedus2021SwitchTS}.

Despite the success of such practice, it remains unclear on the impact of neglecting $\nabla_0$, how to conduct training with only part of the gradient, and whether gradient descent is still effective after neglecting $\nabla_0$.
In this study, we aim to 
% answer the following questions: \emph{How $\appnabla_{\mbox{\scriptsize ST}}$ approximates $\nabla$ and how it can be improved?}
\emph{bridge backpropagation and expert routing by providing a scalable and reliable approximation of $\nabla_0$ and scale MoE models without 
% makes it possible to scal, thereby ensuring
$\nabla_0$ being neglected}.  
% \section{ST Revisited and Scalable Gradient Approximation}
\section{From Discrete to Sparse: \ours}
\label{sec:sparse}
% Here, we first revisit the derivations for positioning $\appnabla_{\mbox{\scriptsize ST}}$ as a first-order approximation. 
Although ST estimators bridged discrete variables and backpropagation, they require the network to be densely activated. 
Here, we first discuss the intrinsic limitation of ST estimators.
Then, we go beyond discrete and bridge sparse expert routing and backpropagation. 

\subsection{Why Existing ST Estimators are not Scalable?}
Targeting to approximate gradients for discrete variables in the general multinomial case, we formally establishes that $E[\appnabla_{\mbox{\scriptsize ST}}]$ is a first-order approximation of $\nabla_0$ in \citet{Liu2023BridgingDA}. 
To discuss ST in the general setting, we reparameterize the expert network $\vy \gets \vpi_\mD f_\mD$ as a function of discrete variables, marked as $\vy \gets h(\mD) = \sum_i \mD_i \vpi_{\mI_i} f_{\mI_i}$. 
Then, we have\footnote{Commonly referred to as baseline subtraction. Note $\sum_i E[g] \frac{\partial\,\vpi_{\mI_i}}{\partial\,W_r} = E[g] \frac{\partial\,\sum_{\mI_i} \vpi_{\mI_i}}{\partial\,W_r}=E[g] \frac{\partial\,\bm{1}}{\partial\,W_r}=0$.}:
\begin{eqnarray}
    \nabla_0 =  \sum_{\mI_i} (h(\mI_i) - E[h]) \frac{\partial\,\vpi_{\mI_i}}{\partial\,W_r} = \sum_{\mI_i} \sum_{\mI_j}  \vpi_{\mI_j} (h(\mI_i) - h(\mI_j)) \frac{\partial\,\vpi_{\mI_i}}{\partial\,W_r}.
    \label{eqn:st_baseline}
\end{eqnarray}
Specifically, approximating $h(\mI_i) - h(\mI_j)$ as $h'({\mI_j}) \cdot ({\mI_i} - {\mI_j})$, the resulting gradient approximation will have the same form as $E[\appnabla_{\mbox{\scriptsize ST}}]$~\citep{Liu2023BridgingDA}.  
In numerical analyses, this approximation is known as the forward Euler method (briefly introduced in Appendix~\ref{appendix:ode}), which has first-order accuracy. 
% We also 
In \citet{Liu2023BridgingDA}, we also 
explored higher-order ODE solvers to better approximate $h(\mI_i) - h(\mI_j)$. 
However, all these attempts require the network to be densely activated, since $\frac{\partial h}{\partial \mD} = \sum_{\mI_i} \mD_i \vpi_{\mI_i} f_{\mI_i}$ necessitates the computation of $\{f_{\mI_1}, \cdots, f_{\mI_N}\}$. 
In order words, although those ST estimators bridge discrete and backpropagation, their computations are dense instead of sparse, blocking their application on MoE training. 

\subsection{Expert Routing Gradient Approximation: Backpropagation Made Sparse}
To bridge the gap between sparse MoE routing and back-propagation, we need to approximate $\nabla_0$ without requiring outputs from all experts. 
In our study, we present a novel framework to move beyond ST and bridge backpropagation and sparse expert routing. 

\smallsection{Gradient Approximation for Expert Routing}
Here, we start by introducing the most simple gradient estimator, i.e., $\appnabla_{\mbox{\scriptsize \ours-1st}}$, where
\begin{eqnarray*}
\appnabla_{\mbox{\scriptsize \ours-1st}} := \frac{\partial g(\pi_\mD f_\mD(\vx))}{\partial\,W_r}.
\end{eqnarray*}
Similar to $E[\appnabla_{\mbox{\scriptsize ST}}]$, $E[\appnabla_{\mbox{\scriptsize \ours-1st}}]$ is a first-order approximation of $\nabla_0$.
To demonstrate this, we take an alternative approach to rewrite $\nabla_0$:
\begin{eqnarray}
    \nabla_0 =  \sum_{\mI_i} (g(\pi_{\mI_i} f_{\mI_i}) - g(\mZero)) \frac{\partial\,\vpi_{\mI_i}}{\partial\,W_r}.
    \label{eqn:fastmax_baseline}
\end{eqnarray}

Adopting the Euler method, we estimate $g(\vpi_{\mI_i} f_{\mI_i}) - g(\mZero)$ as $g'(\vpi_{\mI_i} f_{\mI_i}) \cdot \vpi_{\mI_i} f_{\mI_i}$. 
Then, we have:
\begin{eqnarray*}
\nabla_0 \overset{\mathrm{forward\,Euler}}{\approx} \sum_{\mI_i} g'(\vpi_{\mI_i} f_{\mI_i}) \cdot \vpi_{\mI_i} f_{\mI_i} \cdot \frac{\partial\,\vpi_{\mI_i}}{\partial\,W_r} = E_{\mD \sim \vpi}[\frac{\partial g(\pi_\mD f_\mD(\vx))}{\partial\,W_r}] = E[\appnabla_{\mbox{\scriptsize \ours-1st}}].
\end{eqnarray*}

\smallsection{Gradient Approximation for General Discrete Variables}
To compare with existing ST estimators, we apply $\appnabla_{\mbox{\scriptsize \ours-1st}}$ 
% let us apply this approximation 
to the general case. 
Similar to Equation~\ref{eqn:fastmax_baseline}, we have 
\begin{eqnarray}
    \nabla_0 =  \sum_{\mI_i} (h(\mI_i) - h(\mZero)) \frac{\partial\,\vpi_{\mI_i}}{\partial\,W_r}  \overset{\mathrm{forward\,Euler}}{\approx} \sum_{\mI_i} h'(\mI_i)\cdot \mI_i \cdot \frac{\partial\,\vpi_{\mI_i}}{\partial\,W_r}.
    \label{eqn:fastmax_baseline_h}
\end{eqnarray}
Notably, the first-order approximation of Equation~\ref{eqn:fastmax_baseline_h} only requires the output of one expert, i.e., 
\begin{eqnarray}
h'({\mI_i}) \cdot {\mI_i} = \sum_{\mI_j} \mD_{\mI_j} \cdot \vpi_{\mI_j} \cdot f_{\mI_j} \cdot \mI_i = \vpi_{\mI_j} f_{\mI_j}.
\end{eqnarray}
In other words, Equation~\ref{eqn:fastmax_baseline_h}, taking $h(\mZero)$ as the baseline, leverages the one-hot representation $\mI_i$ to reduce the computation requirement of unactivated experts, thus achieving sparse computations. 
Meanwhile, the first-order approximation of Equation~\ref{eqn:st_baseline}, i.e., $h'({\mI_j}) \cdot ({\mI_i} - {\mI_j})$, has the term $h'({\mI_j}) \cdot {\mI_i}$ and requires a dense computation.

As a summary, both $\appnabla_{\mbox{\scriptsize ST}}$ and $\appnabla_{\mbox{\scriptsize \ours-1st}}$ adopt the forward Euler method and achieve first-order accuracy. 
At the same time, $\appnabla_{\mbox{\scriptsize \ours-1st}}$ only requires the output of one expert thus not sacrificing scalability, while $\appnabla_{\mbox{\scriptsize ST}}$ requires the output of all experts. 

\subsection{Achieving Second-Order Accuracy with the Mid-point Method}

The literature on numerical methods for differential equations shows that it is possible to achieve higher-order accuracy \emph{without computing higher-order derivatives}.
To furnish accurate gradient approximations, we employ a second-order ODE method, the mid-point method (briefly introduced in Appendix~\ref{appendix:ode}).
Specifically, $\appnabla_{\mbox{\scriptsize \oursrd}}$ is a second-order approximation of $\nabla$, where
\begin{eqnarray*}
\appnabla_{\mbox{\scriptsize \oursrd}} := 2\cdot \frac{\partial g(\frac{\pi_\mD f_\mD(\vx)}{2})}{\partial\,W_r}.
\end{eqnarray*}

To demonstrate the connection between $\appnabla_{\mbox{\scriptsize \oursrd}}$ and the mid-point method, we employ the mid-point method to approximate $g(\pi_{\mI_i} f_{\mI_i}) - g(\mZero)$ as $g'(\frac{\pi_{\mI_i} f_{\mI_i}}{2}) \cdot \pi_{\mI_i} f_{\mI_i}$, which also requires only the output of one expert. 
Similarly, it is easy to note:
\begin{eqnarray*}
\nabla_0 \overset{\mathrm{mid-point}}{\approx} \sum_{\mI_i} g'(\frac{\pi_{\mI_i} f_{\mI_i}}{2}) \cdot \vpi_{\mI_i} f_{\mI_i} \cdot \frac{\partial\,\vpi_{\mI_i}}{\partial\,W_r} = E_{\mD \sim \vpi}[2\cdot \frac{\partial g(\frac{\pi_\mD f_\mD(\vx)}{2})}{\partial\,W_r}] = E[\appnabla_{\mbox{\scriptsize \oursrd}}].
\end{eqnarray*}
% Accordingly, $E[\appnabla_{\mbox{\scriptsize \oursrd}}]$ is a second-order approximation of $\nabla$, where
% \begin{eqnarray*}
% \appnabla_{\mbox{\scriptsize \oursrd}} := 2\cdot \frac{\partial g(\frac{\pi_\mD f_\mD(\vx)}{2})}{\partial\,W_r}.
% \end{eqnarray*}
Notably, it is feasible to employ more advanced ODE solvers like RKF4 and approximate $\nabla_0$ with even higher-order accuracy~\citep{fehlberg1969classical}.
In our experiments, we observe that the mid-point method is accurate enough and decide to stick to the mid-point method for simplicity.

\subsection{Balancing Router Training and Expert Training}
Comparing to $\appnabla_{\mbox{\scriptsize \ours-1st}}$, $\appnabla_{\mbox{\scriptsize \oursrd}}$ provides better gradient estimation for router training.
However, $\appnabla_{\mbox{\scriptsize \oursrd}}$ causes additional difficulties for expert training.
Specifically, $\appnabla_{\mbox{\scriptsize \oursrd}}$ requires to change the MoE output from $\vy \gets \pi_\mD f_\mD(\vx)$ to $\vy \gets \frac{\pi_\mD f_\mD(\vx)}{2}$, leading to a gap between the training ($\vy \gets \frac{\pi_\mD f_\mD(\vx)}{2}$) and the inference ($\vy \gets \pi_\mD f_\mD(\vx)$). 
As discussed in Section~\ref{subsec:ablation}, such gap 
% $\appnabla_{\mbox{\scriptsize \oursrd}}$ 
creates significant obstacles for MoE training. 

Meanwhile, $\mD$ is assigned as  $\mD \gets \argmax_{\mI_i} \vpi_{\mI_i}$ during the inference, instead of being sampled from $\vpi$. 
% Thus, it would be sufficient to close the gap by only applying $\appnabla_{\mbox{\scriptsize \oursst}}$ when $\mD$ equals to $\argmax_{\mI_i} \vpi_{\mI_i}$. 
Thus, it would be sufficient to close the gap by only applying $\appnabla_{\mbox{\scriptsize \oursrd}}$ when $\mD \neq \argmax_{\mI_i} \vpi_{\mI_i}$. 
Accordingly, we propose \ours to balance router training and expert training:
\begin{eqnarray*}
\appnabla_{\mbox{\scriptsize \ours}} := (1 - \delta_{\mD}) \appnabla_{\mbox{\scriptsize \oursrd}} + \delta_{\mD} \appnabla_{\mbox{\scriptsize \oursst}}, \mbox{where }  
\delta_{\mD} = 
\begin{cases}
    1,  & \text{if } \mD = \underset{\mI_i}{\argmax}\, \vpi_{\mI_i} \\
    0,  & \text{otherwise} 
\end{cases}.
\end{eqnarray*}

\smallsection{Computational Efficiency of \ours}
$\appnabla_{\mbox{\scriptsize \ours}}$ does not require Hessian or other second-order derivatives, thus having negligible computation overheads (empirical verifications are discussed in Section~\ref{subsec:efficiency}).
At the same time, similar to $\appnabla_{\mbox{\scriptsize ST}}$, our proposed algorithm can be easily integrated with popular library
% automatic differentiation toolkits 
like PyTorch, making it easy to be integrated with existing algorithms. 

\subsection{From Simplified MoE to Switch Transformer}
\label{subsec:setting}

As mentioned in Section~\ref{sect:notation}, our modeling of MoE is a simplified Switch Transformer. 
Here, we first discuss the difference between our simplified setting and Switch Transformer, and then move to necessary modifications to apply \ours to Switch Transformer.

% \smallsection{Scaling Factor}
% In \citet{Fedus2021SwitchTS}, the 

% \smallsection{Sampling of $\mD$}
\smallsection{Discussion on Setting Difference}
The difference between our simplified setting and switch Transformer is the sampling of $\mD$. 
Specifically, in our simplified setting, we assume $\mD$ is sampled from $\vpi$; in 
Switch Transformer, $\mD$ is sampled as:
\begin{eqnarray}
\label{eqn:switch_sampling}
\mD = \argmax_{\mI_i} (\vtheta_{\mI_i} \cdot u_{\mI_i}), \;\mbox{ where }\; u_{\mI_i} \overset{\mathrm{iid}}{\sim} \mbox{Uniform}(1 - r, 1 + r).
\end{eqnarray}

As discussed in \citet{Fedus2021SwitchTS}, directly sampling $\mD$ from $\vpi$ leads to notable performance degradation (also discussed in Section~\ref{subsec:discussion}). 
Meanwhile, in the Switch Transformer, the distribution of $\mD$ does have no analytical form and thus no analytical gradients, making \ours not directly applicable. 
In our experiments, we deploy a sampling process that is differentiable as sampling from $\vpi$, while sharing some important properties with Switch Transformer

\smallsection{Sampling Property of Switch Transformer}
Here, we mark  $\vtheta^* := \max_{\mI_i} \vtheta_{\mI_i}$. 
Then, in Switch Transformer, $\mI_i$ will never be sampled if $\vtheta^* - \vtheta_{\mI_i} > r \cdot (|\vtheta^*| + |\vtheta_{\mI_i}|)$. 
In other words, the distribution of $\mD$ in switch Transformer is masked: small probabilities would directly drop to zero once the corresponding logits hit a threshold. 
In our experiments, we observe that such sparse distribution plays a crucial role in the success of MoE (as elaborated in Section~\ref{subsec:discussion}). 

\smallsection{Applying \ours to Switch Transformer}
Correspondingly, we changed the computation of $\vpi$ from $\vpi_i = \mbox{\small softmax}(\vtheta)_i = \frac{\exp(\vtheta_i)}{\sum_{j=1}^n \exp(\vtheta_j)}$, to $\vpi_i = \frac{\exp(\vtheta_i)\cdot \Delta_{i}}{\sum_{j=1}^n \exp(\vtheta_j)\cdot \Delta_{j}}$, where $\Delta_{j} = \delta(\vtheta^* - \vtheta_{\mI_i} \leq r \cdot (|\vtheta^*| + |\vtheta_{\mI_i}|))$.
In other words, we apply a mask to the softmax function, in order to sample only from experts that are not masked by the Switch Transformer. 

Additionally, since the value of $\vpi$ will be different after applying the mask (which impacts the gradient magnitude of other components), we further changed the output of the MoE layer from $\vpi_{\mD} \cdot f_{\mD}(\vx)$ to $\vomega \cdot \vpi_{\mD} \cdot f_{\mD}(\vx)$, where $\vomega$ is 
% a vector of the word embedding dimension and 
 trainable and is initialized as the $\mOne$ vector. 
Intuitively, $\vomega$ can be viewed as an adaptation on the learning rate for training expert networks. 
Note that, $\vomega$ can be re-parameterized into the feedforward layer after training. 

\section{Experiments}
% Here, we conduct experiments on Neural Machine Translation and Language Model Pretraining. 

Here, we conduct experiments on both pretraining and neural machine translation tasks. 
We closely follow the experiment setting of the existing study. 
Due to the constraint of computation resources, we left MoE related hyper-parameters untuned in all settings, i.e., the jitter (i.e., $r$ in Equation~\ref{eqn:switch_sampling}) is set to 0.1 and the ratio for the load balancing loss is set to 0.01~\citep{Fedus2021SwitchTS}. 
Detailed experiment configurations are elaborated in Appendix~\ref{appendix:expsetting}.

\begin{table}[b!]

\centering
\caption{BLEU score on WMT’14 En-De ($N$ refers to the number of experts). }
\label{table:wmt14}

\begin{tabular}{l|c|ccccc}
\toprule
% {\small$|\mbox{Latent Space}|$}  &  & $4096$ & \multicolumn{4}{c|}{$2^{48}$} & $10^{30}$ \\ 
% \midrule
  & \multirow{2}{*}{Dense} & \multicolumn{5}{c}{Mixture-of-Expert}\\
  &  & $N=2$ & $N=4$ & $N=6$ & $N=8$ & $N=16$ \\ 
\midrule
Transformer-base    & 28.33 & / & / &  / & / & / \\ 
Switch  & / & 28.17 & 28.05 & 27.96 & 27.99 & 27.81 \\ 
Switch+\ours & / & \textbf{28.72} & \textbf{28.61} & \textbf{28.32} & \textbf{28.12} & \textbf{28.08} \\ 
\bottomrule
\end{tabular}
\end{table}

% \begin{table*}[t]
% \centering
% \caption{Results on the GLUE development set. S refers to Switch and S+S refers to Switch+SparseMixer. AVG is the average score across eight tasks. }
% \label{table:glue}
% \resizebox{\linewidth}{!}{
% \begin{tabular}{l l | cccc cccc | c}
% \toprule
% \multirow{2}{*}{$N$} & \multirow{2}{*}{\textbf{Model}} & \textbf{MNLI-(m/mm)} & \textbf{QQP} & \textbf{QNLI} & \textbf{SST-2} & \textbf{CoLA} & \textbf{RTE} & \textbf{MRPC} & \textbf{STS-B} & \multirow{2}{*}{\textbf{AVG}} \\
% & & \textbf{(Acc.)} & \textbf{(Acc.)} & \textbf{(Acc.)} & \textbf{(Acc.)} & \textbf{(Mat. Corr.)}  & \textbf{(Acc.)} & \textbf{(Acc.)} & \textbf{(Spear. Corr.)} \\
% \midrule
% 1 & {Dense} & 88.72/88.40 & 91.90 & 93.36 & 93.35 & 68.71 & 82.31 & 89.95 & 90.83 & 87.37 \\
% \midrule
% % \multicolumn{10}{c}{\textbf{Switch} ($N$ refers to the number of experts)}\\
% % \midrule 
% \multirow{2}{*}{2} & S & 88.55/88.34 & 91.86 & 93.52 & 94.27 & 67.90 & 83.76 & 90.69 & 90.52 & 87.62\\
% & S+S & 89.06/88.78 & 91.98 & 93.54 & 94.38 & 69.96 & 85.20 & 91.67 & 90.81 & 88.31 \\
% \midrule 
% \multirow{2}{*}{4}  & S & 88.12/88.40 & 91.73 & 93.21 & 93.92 & 70.89 & 77.26 & 90.44 & 90.49 & 87.02 \\
% & S+S & 88.97/88.41 & 91.92 & 93.54 & 94.04 & 71.00 & 80.87 & 90.69 & 90.72 & 87.63 \\
% \midrule 
% \multirow{2}{*}{8}  & S & 88.43/88.22 & 91.78 & 93.23 & 94.84 & 68.06 & 80.87 & 90.44 & 90.62 & 87.27 \\
%   & S+S & 88.69/88.47 & 92.03 & 93.41 & 94.15 & 69.00 & 83.76 & 89.95 & 90.81 & 87.71 \\
% \bottomrule
% \end{tabular}
% }
% \end{table*}

\begin{table*}[t]
\centering
\caption{Results on the GLUE development set. S refers to Switch and S+S refers to Switch+SparseMixer. AVG is the average score across eight tasks. }
\label{table:glue}
\resizebox{\linewidth}{!}{
\begin{tabular}{l l | c | cccc cccc}
\toprule
\multirow{2}{*}{$N$} & \multirow{2}{*}{\textbf{Model}}& \multirow{2}{*}{\textbf{AVG}} &  \textbf{MNLI-(m/mm)} & \textbf{QQP} & \textbf{QNLI} & \textbf{SST-2} & \textbf{CoLA} & \textbf{RTE} & \textbf{MRPC} & \textbf{STS-B}  \\
& & & \textbf{(Acc.)} & \textbf{(Acc.)} & \textbf{(Acc.)} & \textbf{(Acc.)} & \textbf{(Mat. Corr.)}  & \textbf{(Acc.)} & \textbf{(Acc.)} & \textbf{(Spear. Corr.)} \\
\midrule
1 & {Dense}& 87.37 & 88.72/88.40 & 91.90 & 93.36 & 93.35 & 68.71 & 82.31 & 89.95 & 90.83  \\
\midrule
% \multicolumn{10}{c}{\textbf{Switch} ($N$ refers to the number of experts)}\\
% \midrule 
\multirow{2}{*}{2} & S & 87.62 &88.55/88.34 & 91.86 & 93.52 & 94.27 & 67.90 & 83.76 & 90.69 & 90.52 \\
& S+S & 88.31 & 89.06/88.78 & 91.98 & 93.54 & 94.38 & 69.96 & 85.20 & 91.67 & 90.81 \\
\midrule 
\multirow{2}{*}{4}  & S& 87.02 & 88.12/88.40 & 91.73 & 93.21 & 93.92 & 70.89 & 77.26 & 90.44 & 90.49  \\
& S+S & 87.63 & 88.97/88.41 & 91.92 & 93.54 & 94.04 & 71.00 & 80.87 & 90.69 & 90.72 \\
\midrule 
\multirow{2}{*}{8}  & S & 87.27 & 88.43/88.22 & 91.78 & 93.23 & 94.84 & 68.06 & 80.87 & 90.44 & 90.62 \\
  & S+S & 87.71 & 88.69/88.47 & 92.03 & 93.41 & 94.15 & 69.00 & 83.76 & 89.95 & 90.81  \\
\bottomrule
\end{tabular}
}
\end{table*}

\begin{figure}[t]
    \centering 
    % \vspace{-2.5mm}
    \includegraphics[width=1.0\textwidth]{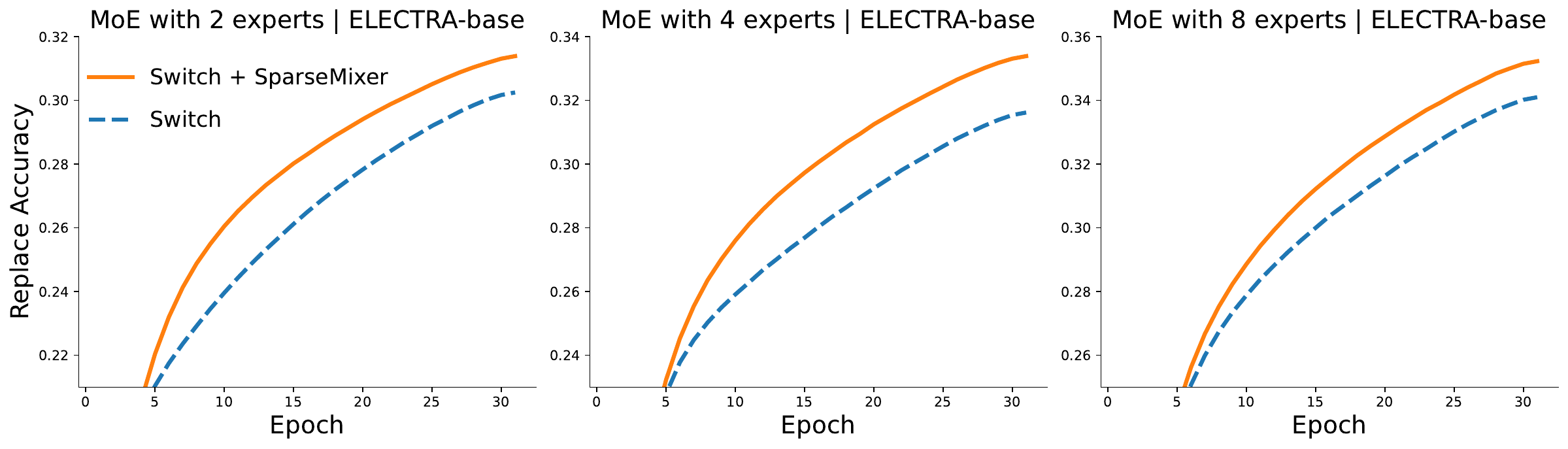}
    % \vspace{-6.0mm}
    \caption{
    Training curves of Switch Transformer on ELECTRA-base training.
    }
    % \vspace{-4mm}
    \label{fig:electra}
\end{figure}

\subsection{Applying \ours on Switch Transformer}

\smallsection{NMT on WMT'14 En-De}
We visualized the training curve in Figure~\ref{fig:wmt14} and summarized the BLEU score in Table~\ref{table:wmt14}. 
Regarding both convergence speed and the final performance, Switch+\ours consistently outperforms Switch in all five settings.
Notably, Switch+\ours matches the training performance of Switch with about \emph{50\% less training updates when $N \in \{4, 6, 8\}$} and about \emph{40\% less training updates when $N \in \{2, 16\}$}. 

% Similar to \citet{Zuo2021TamingSA}, we observe that MoE models are prone to overfitting. 
We can observe that, with more experts, MoE models achieve lower training loss with a worse BLEU score.
Specifically, although Switch Transformer achieves better training performance, its final performance (BLEU score) never outperforms the Dense model, regardless of how many experts it has. 
We believe it requires more data to fully unleash the potential of MoE and suggest this phenomenon indicates that MoE models are prone to overfitting~\citep{Zuo2021TamingSA}. 
% Yet, and would leave in-depth analyses on the performance degeneration to future work.

% Yet, as model architecture design and scaling strategy are beyond the scope of this study, we would conduct more in-depth analyses in future work. 

% Also, it is worth mentioning that, although Switch Transformer achieves better training performance compared to the Dense model, its final performance (BLEU score) never outperforms the Dense model, regardless of how many experts it have. 
% This observation matches the literature that MoE models are prune to overfitting. 

Meanwhile, without changing hyper-parameters or model architectures, the downstream performance of Switch + \ours outperforms both Dense and Switch, when 
% the number of experts is set to
$N \in \{2, 4\}$. 
Specifically, \ours improves the performance of Switch from 28.17 to 28.72 (when $N=2$) and from 28.05 to 28.61 (when $N=4$). 
This phenomenon implies that, with the help of SparseMixer, a sound gradient estimator, MoE learns an expert routing that generalizes better. 

\smallsection{Pretraining}
Following previous work~\citep{Dong2023UnderstandAM}, we visualized the training curve in Figure~\ref{fig:electra} and summarized the fine-tuning results in Table~\ref{table:glue}. 
Regarding both convergence speed and downstream performance, Switch+SparseMixer consistently outperforms Switch in all settings. 
Also, similar to the experiments on machine translation, we observe that MoE models are easier to overfit and both settings achieve the best downstream performance with two experts. 

Also, it is worth mentioning that, while Switch Transformer only outperforms the dense model when the number of experts is set to 2, Switch + SparseMixer consistently outperforms the Dense model in all four settings. 
This phenomenon further verifies our intuition that \ours facilitates MoE models with better expert router training, thus having the resulting model to generalize better.  

\begin{figure}[t]
    \centering 
    % \vspace{-2.5mm}
    \includegraphics[width=1.0\textwidth]{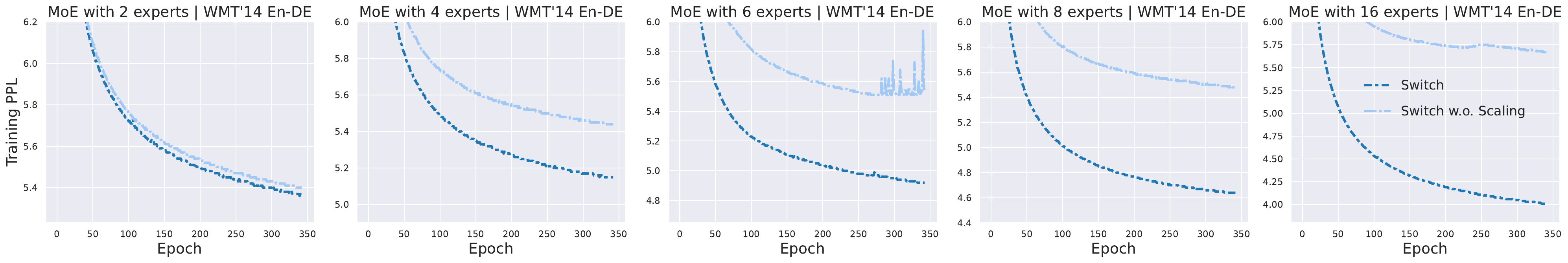}
    % \vspace{-4.0mm}
    \caption{
    Comparison between \emph{Switch Transformer} and \emph{Switch Transformer without Scaling}.
    }
    % \vspace{-2mm}
    \label{fig:switch_ablation}
\end{figure}

\subsection{Discussions}
\label{subsec:discussion}

Here, we conduct experiments to discuss our modeling of the MoE layer as in Section~\ref{sect:notation}. 

\smallsection{Importance of Scaling Expert Outputs with Gating Networks}
One important design detail of MoE is to scale the output of the expert network with the gating network. 
Specifically, the output of the MoE layer is computed as $\vy \gets \vpi_\mD f_\mD(\vx)$, instead of $\vy \gets f_\mD(\vx)$.
This scaling design greatly facilitates the derivation of SparseMixer in Section~\ref{sec:sparse}, and inspires the introduction of $\vomega$ (further discussed in Section~\ref{subsec:ablation}).
Here, we empirically demonstrate that this scaling design also plays an important role in Switch Transformer. 

Specifically, we conduct experiments with a variant of Switch Transformer, i.e., Switch w.o. Scaling, which sets the output of the MoE layer as $\vy \gets f_\mD(\vx)$. 
We apply this Switch variant on WMT'14 En-De and visualize the training curve in Figure~\ref{fig:switch_ablation}. 
Switch ($\vy \gets \vpi_\mD f_\mD(\vx)$) significantly outperforms this variant ($\vy \gets f_\mD(\vx)$). 
Also, we can observe that, when the number of experts is set to 6, using this variant would lead to additional training instability, which further demonstrates the importance of the scaling design.

\begin{figure}[t]
    \centering 
    % \vspace{-2.5mm}
    \includegraphics[width=1.0\textwidth]{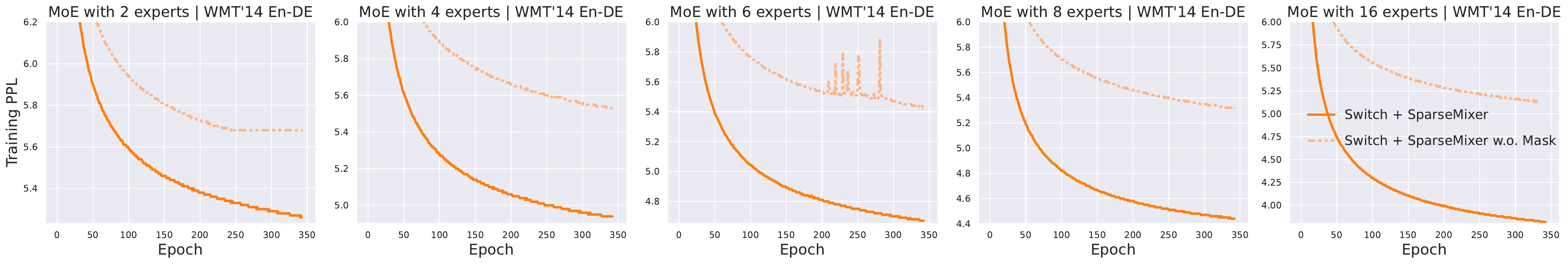}
    % \vspace{-4.0mm}
    \caption{
    Comparison between \emph{\ours} and \emph{\ours without applying mask to sampling}.
    }
    % \vspace{-2mm}
    \label{fig:sparsemixer_ablation}
\end{figure}
\smallsection{Importance of Applying Mask to Softmax}
In Section~\ref{subsec:setting}, we identify that the sampling in Switch Transformer plays an important role in the success of Switch Transformer. 
As discussed in \citet{Fedus2021SwitchTS}, directly using softmax sampling would lead to an inferior performance.

Here, we demonstrate that this masked softmax sampling also plays an important role in Switch + SparseMixer. 
Specifically, we conduct experiments with a variant of SparseMixer, i.e., \ours w.o. Mask, which computes $\vpi_i \gets \mbox{softmax}(\vtheta)_i$. 
We apply \ours w.o. Mask on WMT'14 En-De and visualize the training curve in Figure~\ref{fig:sparsemixer_ablation}. 
\ours ($\vpi_i \gets \frac{\exp(\vtheta_i)\cdot \Delta_{i}}{\sum_{j=1}^n \exp(\vtheta_j)\cdot \Delta_{j}}$) significantly outperforms this variant ($\vy \gets \mbox{softmax}(\vtheta)_i$). 
Also, we can observe that, when the number of experts is set to 6, using this variant would lead to additional training instability, which further demonstrates the importance of applying mask to softmax. 

\begin{figure}[t]
    \centering 
    % \vspace{-2.5mm}
    \includegraphics[width=1.0\textwidth]{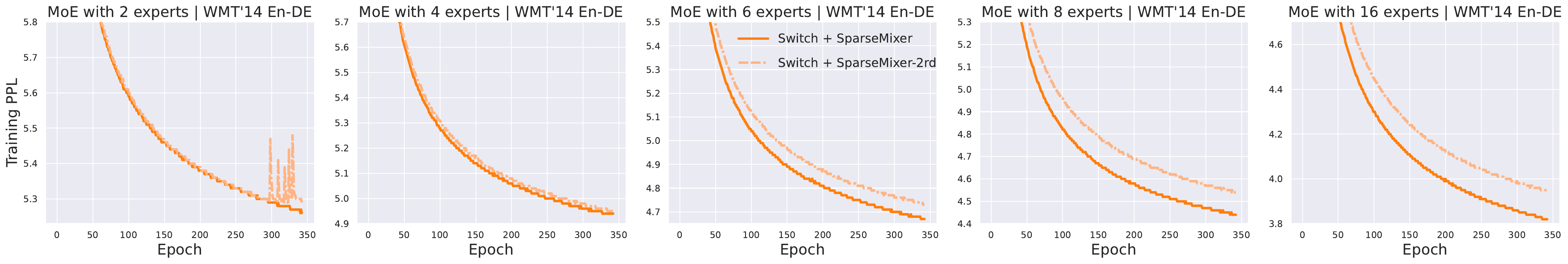}
    % \vspace{-4.0mm}
    \caption{
    Comparison between \emph{\ours} and \emph{\oursrd}.
    }
    % \vspace{-2mm}
    \label{fig:amp_ablation}
\end{figure}

\subsection{Ablation}
\label{subsec:ablation}

Here, we conduct experiments to discuss the design details of SparseMixer. 

\smallsection{Importance of Balancing Expert Learning and Routing Learning}
While \oursrd provides better gradient approximation for expert routing, it creates a gap between training and inference. 
To demonstrate the importance of balancing router training and expert training, we conduct experiments on applying \oursrd on WMT'14 En-De. 
As visualized in Figure~\ref{fig:amp_ablation}, \ours consistently outperforms \oursrd in all cases. 
Also, \oursrd exhibits training instability when setting the number of experts to 2. 

\smallsection{Mid-point Method and $\vomega$ Scaling}
To better understand the benefit introducing $\vomega$ (as in Section~\ref{subsec:setting}) and make comparisons with SparseMixer-1st, we conduct additional ablation studies on WMT'14 En-De. 
Specifically, we consider two \ours variants:
\begin{itemize}[leftmargin=*]
    \item
    ablation-1 removes $\vomega$ from \ours (i.e., changes the output of Switch + SparseMixer from $\vomega \cdot \vpi_\mD \cdot f_\mD(\vx)$ to $\vpi_\mD \cdot f_\mD(\vx)$). 
    
    \item 
    ablation-2 further replaces the mid-point method with the forward Euler method in SparseMixer-ablation-1, i.e., $\appnabla_{\mbox{\scriptsize \oursst}} $ is employed as the gradient estimator and $\vomega$ is removed.
\end{itemize}
We apply these two variants to WMT'14 En-De and visualize their training curve in Figure~\ref{fig:wmt14}. 
The results further verified our intuition that $\vomega$ facilitates MoE training by alleviating the impact of applying masks.
Also, it shows that integrating the mid-point method helps to better approximate expert routing gradient.

\begin{table}[t]

\centering
\caption{Average Training Time Cost (s/update). $N$ refers to the number of experts.}
\label{table:efficiency}
% \vspace{.2cm}
\scalebox{.9}{
\begin{tabular}{l|ccccc|ccc}
\toprule
  & \multicolumn{5}{c|}{WMT’14 En-De} & \multicolumn{3}{c}{Pretraining}    \\   
  & $N=2$ & $N=4$ & $N=6$ & $N=8$ & $N=16$ &  $N=2$ & $N=4$ & $N=8$  \\ 
  \midrule
Switch & 0.32 & 0.33 & 0.34 & 0.36 & 0.40 & 1.87 & 1.90 & 1.98 \\
  \midrule
Switch + \ours & 0.32 & 0.33 & 0.34 & 0.36 & 0.40 & 1.87 & 1.90 & 1.98  \\
\bottomrule
\end{tabular}
}
% \vspace{-.3cm}
\end{table}

\subsection{Efficiency}
\label{subsec:efficiency}
We summarized the average time cost per update in Table~\ref{table:efficiency}. 
Switch+\ours achieves an identical average time cost with Switch in all eight settings. 
This shows that the computation overheads of \ours are negligible. 

\section{Conclusion}

In this study, we present \ours to move beyond discrete and bridge the gap between sparse MoE routing and backpropagation. 
Rooted in a numerical ODE framework, \ours harnesses the mid-point method, a second-order ODE solver, to deliver precise gradient approximations with negligible computational overhead. 
In our experiments on both neural machine translation and pretraining tasks, 
\ours not only accelerates training convergence by up to two times but also facilitates MoE with properly trained expert routing. 
Remarkably, while Switch Transformer underperforms the dense model in all three pretraining settings, incorporating \ours as the gradient estimator allows the resulting MoE models to consistently outperform the dense model. 

There are multiple interesting directions to be explored in the future. 
While our method is based on first-order and second-order ODE solvers, it would be interesting to apply higher-order ODE solvers and even adaptive ODE solvers like RKF4~\citep{fehlberg1969classical}. 
Also, since our study paves the way towards designing gradient approximation for scalable MOE training, we plan to further improve the architecture design of MoE models.
In the end, as we observed that MoE models are easier to overfit, we plan to study the scaling law of sparse models and facilitate large-scale pre-training.
% , under the guidance of SparseMixer. 

% \clearpage
\bibliography{iclr2024_conference}

\begin{thebibliography}{41}
\providecommand{\natexlab}[1]{#1}
\providecommand{\url}[1]{\texttt{#1}}
\expandafter\ifx\csname urlstyle\endcsname\relax
  \providecommand{\doi}[1]{doi: #1}\else
  \providecommand{\doi}{doi: \begingroup \urlstyle{rm}\Url}\fi

\bibitem[Ascher \& Petzold(1998)Ascher and Petzold]{Ascher1998ComputerMF}
Uri~M. Ascher and Linda~R. Petzold.
\newblock \emph{Computer methods for ordinary differential equations and differential-algebraic equations}.
\newblock 1998.

\bibitem[Bajaj et~al.(2022)Bajaj, Xiong, Ke, Liu, He, Tiwary, Liu, Bennett, Song, and Gao]{Bajaj2022METROED}
Payal Bajaj, Chenyan Xiong, Guolin Ke, Xiaodong Liu, Di~He, Saurabh Tiwary, Tie-Yan Liu, Paul Bennett, Xia Song, and Jianfeng Gao.
\newblock Metro: Efficient denoising pretraining of large scale autoencoding language models with model generated signals.
\newblock \emph{ArXiv}, abs/2204.06644, 2022.

\bibitem[Bengio et~al.(2013)Bengio, L{\'e}onard, and Courville]{Bengio2013EstimatingOP}
Yoshua Bengio, Nicholas L{\'e}onard, and Aaron~C. Courville.
\newblock Estimating or propagating gradients through stochastic neurons for conditional computation.
\newblock \emph{ArXiv}, abs/1308.3432, 2013.

\bibitem[Bojar et~al.(2014)Bojar, Buck, Federmann, Haddow, Koehn, Leveling, Monz, Pecina, Post, Saint-Amand, et~al.]{bojar2014findings}
Ond{\v{r}}ej Bojar, Christian Buck, Christian Federmann, Barry Haddow, Philipp Koehn, Johannes Leveling, Christof Monz, Pavel Pecina, Matt Post, Herve Saint-Amand, et~al.
\newblock Findings of the 2014 workshop on statistical machine translation.
\newblock In \emph{Workshop on Statistical Machine Translation}, 2014.

\bibitem[Chowdhery et~al.(2022)Chowdhery, Narang, Devlin, Bosma, Mishra, Roberts, Barham, Chung, Sutton, Gehrmann, Schuh, Shi, Tsvyashchenko, Maynez, Rao, Barnes, Tay, Shazeer, Prabhakaran, Reif, Du, Hutchinson, Pope, Bradbury, Austin, Isard, Gur-Ari, Yin, Duke, Levskaya, Ghemawat, Dev, Michalewski, Garcia, Misra, Robinson, Fedus, Zhou, Ippolito, Luan, Lim, Zoph, Spiridonov, Sepassi, Dohan, Agrawal, Omernick, Dai, Pillai, Pellat, Lewkowycz, Moreira, Child, Polozov, Lee, Zhou, Wang, Saeta, Diaz, Firat, Catasta, Wei, Meier-Hellstern, Eck, Dean, Petrov, and Fiedel]{chowdhery2022palm}
Aakanksha Chowdhery, Sharan Narang, Jacob Devlin, Maarten Bosma, Gaurav Mishra, Adam Roberts, Paul Barham, Hyung~Won Chung, Charles Sutton, Sebastian Gehrmann, Parker Schuh, Kensen Shi, Sasha Tsvyashchenko, Joshua Maynez, Abhishek Rao, Parker Barnes, Yi~Tay, Noam Shazeer, Vinodkumar Prabhakaran, Emily Reif, Nan Du, Ben Hutchinson, Reiner Pope, James Bradbury, Jacob Austin, Michael Isard, Guy Gur-Ari, Pengcheng Yin, Toju Duke, Anselm Levskaya, Sanjay Ghemawat, Sunipa Dev, Henryk Michalewski, Xavier Garcia, Vedant Misra, Kevin Robinson, Liam Fedus, Denny Zhou, Daphne Ippolito, David Luan, Hyeontaek Lim, Barret Zoph, Alexander Spiridonov, Ryan Sepassi, David Dohan, Shivani Agrawal, Mark Omernick, Andrew~M. Dai, Thanumalayan~Sankaranarayana Pillai, Marie Pellat, Aitor Lewkowycz, Erica Moreira, Rewon Child, Oleksandr Polozov, Katherine Lee, Zongwei Zhou, Xuezhi Wang, Brennan Saeta, Mark Diaz, Orhan Firat, Michele Catasta, Jason Wei, Kathy Meier-Hellstern, Douglas Eck, Jeff Dean, Slav Petrov, and Noah Fiedel.
\newblock Palm: Scaling language modeling with pathways.
\newblock \emph{ArXiv}, abs/2204.02311, 2022.

\bibitem[Clark et~al.(2020)Clark, Luong, Le, and Manning]{Clark2020ELECTRA}
Kevin Clark, Minh-Thang Luong, Quoc~V. Le, and Christopher~D. Manning.
\newblock Electra: Pre-training text encoders as discriminators rather than generators.
\newblock In \emph{ICLR}, 2020.

\bibitem[Devlin et~al.(2019)Devlin, Chang, Lee, and Toutanova]{Devlin2019BERTPO}
Jacob Devlin, Ming-Wei Chang, Kenton Lee, and Kristina Toutanova.
\newblock Bert: Pre-training of deep bidirectional transformers for language understanding.
\newblock In \emph{NAACL}, 2019.

\bibitem[Dong et~al.(2023)Dong, Liu, Cheng, Shang, Gao, and Liu]{Dong2023UnderstandAM}
Chengyu Dong, Liyuan Liu, Hao Cheng, Jingbo Shang, Jianfeng Gao, and Xiaodong Liu.
\newblock Understand and modularize generator optimization in electra-style pretraining.
\newblock In \emph{ICML}, 2023.

\bibitem[Fedus et~al.(2021)Fedus, Zoph, and Shazeer]{Fedus2021SwitchTS}
William Fedus, Barret Zoph, and Noam~M. Shazeer.
\newblock Switch transformers: Scaling to trillion parameter models with simple and efficient sparsity.
\newblock \emph{ArXiv}, abs/2101.03961, 2021.

\bibitem[Fehlberg(1969)]{fehlberg1969classical}
Erwin Fehlberg.
\newblock Classical fifth-and seventh-order runge-kutta formulas with stepsize control.
\newblock \emph{Computing}, 1969.

\bibitem[Grathwohl et~al.(2018)Grathwohl, Choi, Wu, Roeder, and Duvenaud]{Grathwohl2017BackpropagationTT}
Will Grathwohl, Dami Choi, Yuhuai Wu, Geoffrey Roeder, and David~Kristjanson Duvenaud.
\newblock Backpropagation through the void: Optimizing control variates for black-box gradient estimation.
\newblock In \emph{ICLR}, 2018.

\bibitem[Gu et~al.(2016)Gu, Levine, Sutskever, and Mnih]{Gu2015MuPropUB}
Shixiang~Shane Gu, Sergey Levine, Ilya Sutskever, and Andriy Mnih.
\newblock Muprop: Unbiased backpropagation for stochastic neural networks.
\newblock In \emph{ICLR}, 2016.

\bibitem[Gumbel(1954)]{Gumbel1954StatisticalTO}
Emil~Julius Gumbel.
\newblock \emph{Statistical Theory of Extreme Values and Some Practical Applications : A Series of Lectures}.
\newblock 1954.

\bibitem[He et~al.(2020)He, Liu, Gao, and Chen]{He2020DeBERTaDB}
Pengcheng He, Xiaodong Liu, Jianfeng Gao, and Weizhu Chen.
\newblock Deberta: Decoding-enhanced bert with disentangled attention.
\newblock \emph{ArXiv}, abs/2006.03654, 2020.

\bibitem[Jacobs et~al.(1991)Jacobs, Jordan, Nowlan, and Hinton]{Jacobs1991AdaptiveMO}
Robert~A. Jacobs, Michael~I. Jordan, Steven~J. Nowlan, and Geoffrey~E. Hinton.
\newblock Adaptive mixtures of local experts.
\newblock \emph{Neural Computation}, 3:\penalty0 79--87, 1991.

\bibitem[Jang et~al.(2017)Jang, Gu, and Poole]{Jang2016CategoricalRW}
Eric Jang, Shixiang~Shane Gu, and Ben Poole.
\newblock Categorical reparameterization with gumbel-softmax.
\newblock In \emph{ICLR}, 2017.

\bibitem[Jordan \& Jacobs(1994)Jordan and Jacobs]{Jordan1994HierarchicalMO}
Michael~I. Jordan and Robert~A. Jacobs.
\newblock Hierarchical mixtures of experts and the em algorithm.
\newblock \emph{Neural Computation}, 6:\penalty0 181--214, 1994.

\bibitem[Kool et~al.(2021)Kool, Maddison, and Mnih]{Kool2021UnbiasedGE}
Wouter Kool, Chris~J. Maddison, and Andriy Mnih.
\newblock Unbiased gradient estimation with balanced assignments for mixtures of experts.
\newblock In \emph{I (Still) Can’t Believe It’s Not Better Workshop at NeurIPS 2021}, 2021.

\bibitem[Lepikhin et~al.(2020)Lepikhin, Lee, Xu, Chen, Firat, Huang, Krikun, Shazeer, and Chen]{Lepikhin2020GShardSG}
Dmitry Lepikhin, HyoukJoong Lee, Yuanzhong Xu, Dehao Chen, Orhan Firat, Yanping Huang, Maxim Krikun, Noam~M. Shazeer, and Z.~Chen.
\newblock Gshard: Scaling giant models with conditional computation and automatic sharding.
\newblock \emph{ArXiv}, abs/2006.16668, 2020.

\bibitem[Lewis et~al.(2021)Lewis, Bhosale, Dettmers, Goyal, and Zettlemoyer]{Lewis2021BASELS}
Mike Lewis, Shruti Bhosale, Tim Dettmers, Naman Goyal, and Luke Zettlemoyer.
\newblock Base layers: Simplifying training of large, sparse models.
\newblock In \emph{ICML}, 2021.

\bibitem[Liu et~al.(2020{\natexlab{a}})Liu, Jiang, He, Chen, Liu, Gao, and Han]{Liu2019OnTV}
Liyuan Liu, Haoming Jiang, Pengcheng He, Weizhu Chen, Xiaodong Liu, Jianfeng Gao, and Jiawei Han.
\newblock On the variance of the adaptive learning rate and beyond.
\newblock In \emph{ICLR}, 2020{\natexlab{a}}.

\bibitem[Liu et~al.(2020{\natexlab{b}})Liu, Liu, Gao, Chen, and Han]{Liu2020UnderstandingTD}
Liyuan Liu, Xiaodong Liu, Jianfeng Gao, Weizhu Chen, and Jiawei Han.
\newblock Understanding the difficulty of training transformers.
\newblock In \emph{EMNLP}, 2020{\natexlab{b}}.

\bibitem[Liu et~al.(2023)Liu, Dong, Liu, Yu, and Gao]{Liu2023BridgingDA}
Liyuan Liu, Chengyu Dong, Xiaodong Liu, Bin Yu, and Jianfeng Gao.
\newblock Bridging discrete and backpropagation: Straight-through and beyond.
\newblock \emph{ArXiv}, abs/2304.08612, 2023.

\bibitem[Liu et~al.(2019)Liu, He, Chen, and Gao]{Liu2019MultiTaskDN}
Xiaodong Liu, Pengcheng He, Weizhu Chen, and Jianfeng Gao.
\newblock Multi-task deep neural networks for natural language understanding.
\newblock In \emph{ACL}, 2019.

\bibitem[Lu et~al.(2020)Lu, Li, He, Sun, Dong, Qin, Wang, and Liu]{lu2020understanding}
Yiping Lu, Zhuohan Li, Di~He, Zhiqing Sun, Bin Dong, Tao Qin, Liwei Wang, and Tie-Yan Liu.
\newblock Understanding and improving transformer from a multi-particle dynamic system point of view.
\newblock In \emph{ICLR Workshop DeepDiffEq}, 2020.

\bibitem[Maddison et~al.(2014)Maddison, Tarlow, and Minka]{Maddison2014AS}
Chris~J. Maddison, Daniel Tarlow, and Thomas~P. Minka.
\newblock A* sampling.
\newblock In \emph{NIPS}, 2014.

\bibitem[Mustafa et~al.(2022)Mustafa, Riquelme, Puigcerver, Jenatton, and Houlsby]{mustafa2022multimodal}
Basil Mustafa, Carlos Riquelme, Joan Puigcerver, Rodolphe Jenatton, and Neil Houlsby.
\newblock Multimodal contrastive learning with limoe: the language-image mixture of experts.
\newblock \emph{ArXiv}, abs/2206.02770, 2022.

\bibitem[OpenAI(2023)]{openai2023gpt4}
OpenAI.
\newblock Gpt-4 technical report.
\newblock \emph{ArXiv}, abs/2303.08774, 2023.

\bibitem[Ott et~al.(2019)Ott, Edunov, Baevski, Fan, Gross, Ng, Grangier, and Auli]{ott2019fairseq}
Myle Ott, Sergey Edunov, Alexei Baevski, Angela Fan, Sam Gross, Nathan Ng, David Grangier, and Michael Auli.
\newblock fairseq: A fast, extensible toolkit for sequence modeling.
\newblock In \emph{NAACL-HLT Demonstrations}, 2019.

\bibitem[Raffel et~al.(2019)Raffel, Shazeer, Roberts, Lee, Narang, Matena, Zhou, Li, and Liu]{Raffel2019ExploringTL}
Colin Raffel, Noam~M. Shazeer, Adam Roberts, Katherine Lee, Sharan Narang, Michael Matena, Yanqi Zhou, Wei Li, and Peter~J. Liu.
\newblock Exploring the limits of transfer learning with a unified text-to-text transformer.
\newblock \emph{ArXiv}, abs/1910.10683, 2019.

\bibitem[Riquelme et~al.(2021)Riquelme, Puigcerver, Mustafa, Neumann, Jenatton, Pinto, Keysers, and Houlsby]{riquelme2021scaling}
Carlos Riquelme, Joan Puigcerver, Basil Mustafa, Maxim Neumann, Rodolphe Jenatton, André~Susano Pinto, Daniel Keysers, and Neil Houlsby.
\newblock Scaling vision with sparse mixture of experts.
\newblock In \emph{NeurIPS}, 2021.

\bibitem[Rosenblatt(1957)]{rosenblatt1957perceptron}
Frank Rosenblatt.
\newblock \emph{The perceptron, a perceiving and recognizing automaton Project Para}.
\newblock Cornell Aeronautical Laboratory, 1957.

\bibitem[Shazeer et~al.(2017)Shazeer, Mirhoseini, Maziarz, Davis, Le, Hinton, and Dean]{Shazeer2017OutrageouslyLN}
Noam~M. Shazeer, Azalia Mirhoseini, Krzysztof Maziarz, Andy Davis, Quoc~V. Le, Geoffrey~E. Hinton, and Jeff Dean.
\newblock Outrageously large neural networks: The sparsely-gated mixture-of-experts layer.
\newblock In \emph{ICLR}, 2017.

\bibitem[Shi et~al.(2022)Shi, Zhou, Hwang, Titsias, and Mackey]{shi2022gradient}
Jiaxin Shi, Yuhao Zhou, Jessica Hwang, Michalis Titsias, and Lester Mackey.
\newblock Gradient estimation with discrete stein operators.
\newblock In \emph{NeurIPS}, 2022.

\bibitem[Szegedy et~al.(2016)Szegedy, Vanhoucke, Ioffe, Shlens, and Wojna]{szegedy2016rethinking}
Christian Szegedy, Vincent Vanhoucke, Sergey Ioffe, Jon Shlens, and Zbigniew Wojna.
\newblock Rethinking the inception architecture for computer vision.
\newblock In \emph{CVPR}, 2016.

\bibitem[Touvron et~al.(2023)Touvron, Martin, Stone, Albert, Almahairi, Babaei, Bashlykov, Batra, Bhargava, Bhosale, Bikel, Blecher, Ferrer, Chen, Cucurull, Esiobu, Fernandes, Fu, Fu, Fuller, Gao, Goswami, Goyal, Hartshorn, Hosseini, Hou, Inan, Kardas, Kerkez, Khabsa, Kloumann, Korenev, Koura, Lachaux, Lavril, Lee, Liskovich, Lu, Mao, Martinet, Mihaylov, Mishra, Molybog, Nie, Poulton, Reizenstein, Rungta, Saladi, Schelten, Silva, Smith, Subramanian, Tan, Tang, Taylor, Williams, Kuan, Xu, Yan, Zarov, Zhang, Fan, Kambadur, Narang, Rodriguez, Stojnic, Edunov, and Scialom]{touvron2023llama}
Hugo Touvron, Louis Martin, Kevin Stone, Peter Albert, Amjad Almahairi, Yasmine Babaei, Nikolay Bashlykov, Soumya Batra, Prajjwal Bhargava, Shruti Bhosale, Dan Bikel, Lukas Blecher, Cristian~Canton Ferrer, Moya Chen, Guillem Cucurull, David Esiobu, Jude Fernandes, Jeremy Fu, Wenyin Fu, Brian Fuller, Cynthia Gao, Vedanuj Goswami, Naman Goyal, Anthony Hartshorn, Saghar Hosseini, Rui Hou, Hakan Inan, Marcin Kardas, Viktor Kerkez, Madian Khabsa, Isabel Kloumann, Artem Korenev, Punit~Singh Koura, Marie-Anne Lachaux, Thibaut Lavril, Jenya Lee, Diana Liskovich, Yinghai Lu, Yuning Mao, Xavier Martinet, Todor Mihaylov, Pushkar Mishra, Igor Molybog, Yixin Nie, Andrew Poulton, Jeremy Reizenstein, Rashi Rungta, Kalyan Saladi, Alan Schelten, Ruan Silva, Eric~Michael Smith, Ranjan Subramanian, Xiaoqing~Ellen Tan, Binh Tang, Ross Taylor, Adina Williams, Jian~Xiang Kuan, Puxin Xu, Zheng Yan, Iliyan Zarov, Yuchen Zhang, Angela Fan, Melanie Kambadur, Sharan Narang, Aurelien Rodriguez, Robert Stojnic, Sergey Edunov, and Thomas
  Scialom.
\newblock Llama 2: Open foundation and fine-tuned chat models.
\newblock \emph{ArXiv}, abs/2307.09288, 2023.

\bibitem[Tucker et~al.(2017)Tucker, Mnih, Maddison, Lawson, and Sohl-Dickstein]{Tucker2017REBARLU}
G.~Tucker, Andriy Mnih, Chris~J. Maddison, John Lawson, and Jascha~Narain Sohl-Dickstein.
\newblock Rebar: Low-variance, unbiased gradient estimates for discrete latent variable models.
\newblock In \emph{NIPS}, 2017.

\bibitem[Wang et~al.(2018)Wang, Singh, Michael, Hill, Levy, and Bowman]{Wang2018GLUEAM}
Alex Wang, Amanpreet Singh, Julian Michael, Felix Hill, Omer Levy, and Samuel~R. Bowman.
\newblock Glue: A multi-task benchmark and analysis platform for natural language understanding.
\newblock In \emph{BlackboxNLP@EMNLP}, 2018.

\bibitem[Williams(1992)]{Williams1992SimpleSG}
Ronald~J. Williams.
\newblock Simple statistical gradient-following algorithms for connectionist reinforcement learning.
\newblock \emph{Machine Learning}, 8:\penalty0 229--256, 1992.

\bibitem[Zhu et~al.(2015)Zhu, Kiros, Zemel, Salakhutdinov, Urtasun, Torralba, and Fidler]{Zhu2015AligningBA}
Yukun Zhu, Ryan Kiros, Richard~S. Zemel, Ruslan Salakhutdinov, Raquel Urtasun, Antonio Torralba, and Sanja Fidler.
\newblock Aligning books and movies: Towards story-like visual explanations by watching movies and reading books.
\newblock In \emph{ICCV}, 2015.

\bibitem[Zuo et~al.(2022)Zuo, Liu, Jiao, Kim, Hassan, Zhang, Zhao, and Gao]{Zuo2021TamingSA}
Simiao Zuo, Xiaodong Liu, Jian Jiao, Young~Jin Kim, Hany Hassan, Ruofei Zhang, Tuo Zhao, and Jianfeng Gao.
\newblock Taming sparsely activated transformer with stochastic experts.
\newblock In \emph{ICLR}, 2022.

\end{thebibliography}
\bibliographystyle{iclr2024_conference}

\appendix
\section{Forward Euler Method and Midpoint Method}
\label{appendix:ode}

For simplicity, we consider a simple function $g(x): \mathcal{R} \to \mathcal{R}$ that is three times differentiable on $[t_0, t_1]$. 
Now, we proceed to a simple introduction to approximate $\int_{t_0}^{t_1} g'(x) dx$ with the Forward Euler Method and the Midpoint Method. 
For a detailed introduction to numerical ODE methods, please refer to \citet{Ascher1998ComputerMF}. 

\smallsection{Forward Euler Method}
Here, we approximate $g(t_1)$ with the first-order Taylor expansion, i.e.,  $g(t_1) = g(t_0) + g'(t_0) \cdot (t_1- t_0) + O((t_1 - t_0)^2)$, then we have $\int_{t_0}^{t_1} g'(x) dx \approx g'(t_0) (t_1 - t_0)$. 
Since we used the first-order Taylor expansion, this approximation has first-order accuracy. 

\smallsection{Midpoint Method}
First, we approximate $g(t_1)$ with the second-order Taylor expansion: 
\begin{equation}
g(t_1) = g(t_0) + g'(t_0) \cdot (t_1 - t_0) + \frac{g''(t_0)}{2} \cdot (t_1 - t_0)^2 + O((t_1 - t_0)^3).
\label{eqn:taylor-2nd}
\end{equation}
Then, we show that we can match this approximation by using $g'(\frac{t_1+t_0}{2})$. 
Taylor expanding $g'(\frac{t_1+t_0}{2})$ to the first-order, we have:
\begin{equation}
g'(\frac{t_1+t_0}{2}) = g'(t_0) + g''(t_0) \cdot \frac{t_1 - t_0}{2} + O((t_1 - t_0)^2) 
\nonumber
\end{equation}
Therefore, we have:
\begin{equation}
g(t_0) + g'(\frac{t_1+t_0}{2})  (t_1 - t_0) = g(t_0) + g'(t_0) \cdot (t_1 - t_0) + \frac{g''(t_0)}{2} \cdot (t_1 - t_0)^2 + O((t_1 - t_0)^3).
\nonumber
\end{equation}
It is easy to notice that the right-hand side of the above equation matches the second-order Taylor expansion of $g(t_1)$ as in Equation~\ref{eqn:taylor-2nd}.  
Therefore, the above approximation (i.e., approximating $g(t_1) - g(t_0)$ as $g'(\frac{t_1+t_0}{2})  (t_1 - t_0)$) has second-order accuracy.

\smallsection{Connection to $f(\mI_i) - f(\vzero)$}
By setting $g(x) = f(x \cdot \mI_i)$, we have $g(1) - g(0) = f(\mI_i) - f(\vzero)$. 
Then, it is easy to notice that the forward Euler Method approximates $f(\mI_i) - f(\vzero)$ as $\frac{\partial f(\mI_i)}{\partial \mI_i}\mI_i$ and has first-order accuracy. 
Also, the mid-point method approximates $f(\mI_i) - f(\vzero)$ as $\frac{\partial f(\mI_i/2)}{\partial \mI_i/2} \mI_i$ and has second-order accuracy.

\section{Experiment Setting}
\label{appendix:expsetting}

Here, we conduct experiments on both pretraining and neural machine translation tasks. 
We closely follow the experiment setting of the existing study. 
Due to the constraint of computation resources, we left MoE related hyper-parameters untuned in all settings, i.e., jitter ($r$) is set to 0.1 and load balance loss ratio is set to 0.01~\citep{Fedus2021SwitchTS}. 

\subsection{Neural Machine Translation}

\smallsection{Problem Setting}
Our experiments are based on the fairseq package~\citep{ott2019fairseq}.
As to pre-processing, we follow the public released script from previous work~\citep{lu2020understanding}, and conduct evaluations on the provided `newstest14` file. 
More details can be found in \citet{bojar2014findings}. 

\smallsection{Model Architecture}
As to model specifics, we directly adopt the Transformer-base model on the WMT'14 En-De datasets.
Specifically, we use encoder-decoder Transformer models with 6 encoder layers, 6 decoder layers, 512-dimension word embedding, 8-head attentions, and 2048-dimension feed-forward layers. 
Following \citet{Fedus2021SwitchTS}, we apply MoE layers at every other feed-forward layers, set jitter to 0.1, and configure load balance ratio as $1\cdot10^{-2}$. 
As the number of experts, we consider 5 different settings, i.e., $N\in \{2, 4, 6, 8, 16\}$. 
Label smoothed cross-entropy is used as the objective function with the uncertainty set as $0.1$~\citep{szegedy2016rethinking}. 

\smallsection{Training Settings}
We mostly followed \citep{Liu2019OnTV} for training settings. 
Specifically, we use Adam as the optimizer set $(\beta_1, \beta_2)$ as $(0.9, 0.98)$,  use inverse sqrt learning rate scheduler with a warmup phrase (8000 steps).
All dropout ratios (including activation dropout and attention dropout) are set to 0.1. 
The maximum learning rate is set to $7\cdot10^{-4}$ and the maximum token number per batch is set to $2^{17}$.
We conduct training for $4\cdot 10^{5}$ updates and report the performance of the last checkpoint and the checkpoint with the lowest development loss. 

\subsection{Pre-training}

\smallsection{Pre-training Setup}
We follow the standard settings for training Base models~\citep{Clark2020ELECTRA, Bajaj2022METROED, Dong2023UnderstandAM}, 
Specifically, we employ Wikipedia and BookCorpus~\citep{Zhu2015AligningBA} for pre-training and set the sequence length to $512$, which leads to $16$ GB of texts and $256$M samples. 
We use a cased sentence piece BPE vocabulary of $128$K tokens following~\citet{He2020DeBERTaDB}, and conduct pre-training for $125$K updates with a batch size of $2048$ sentences.
% As to the auxiliary model, it is pre-trained following a standard MLM style.

\smallsection{Model Architecture}
Our main model (discriminator) setting follows the BERT$_{\text{base}}$ architecture~\citep{Devlin2019BERTPO}.
Specifically, the model has 12 layers, 768-dimension embedding, and 12-head attention. 
As to the feed-forward networks, we set the number of hidden state dimensions to 3076. 
Following \citet{Bajaj2022METROED} and \citet{Dong2023UnderstandAM}, we further enhanced the model with the T5 relative position encoding~\citep{Raffel2019ExploringTL} and use $32$ bins. 
We set dropout as $0.1$ and employ Admin~\citep{Liu2020UnderstandingTD} for model initialization to stabilize the training.
Following \citet{Fedus2021SwitchTS}, we apply MoE layers at every other feed-forward layers, set jitter to 0.1, and configure load balance ratio as $1\cdot10^{-2}$. 
As the number of experts, we consider 3 different settings, i.e., $N\in \{2, 4, 8\}$. 
As to the auxiliary model, we follow previous works~\citep{Clark2020ELECTRA, Bajaj2022METROED} to set the size of the auxiliary model (generator) to be $4$ layers.

\smallsection{Optimization}
% We follow previous work~\citep{Clark2020ELECTRA,Bajaj2022METROED} to select the generator size, namely $4$ layers for the Base setting. 
We configure the optimizer as Adam, $(\beta_1, \beta_2)$ as $(0.9, 0.98)$, weight decay as $0.01$, the loss weight as $50$, the peak learning rate as $5e-4$, and the warmup steps as $10$K.

\smallsection{Downstream evaluation setup}
We conduct evaluation on downstream tasks following the setup in previous works~\citep{Bajaj2022METROED}. 
Specifically, we conduct single-task, single-model fine-tuning on the GLUE~\citep{Wang2018GLUEAM} benchmark. 
As summarized in the Appendix (Table~\ref{tbl:glue_description}), GLUE includes 9 subtasks. 
Following \cite{Liu2019MultiTaskDN}, we conduct a grid-search on hyper-parameters and report the best performance for both Switch and Swith + SparseMixer. 
The complete search space is included in Appendix (Table~\ref{table:fine-tune}).

\vspace{1.4cm}
\begin{table*}[h!]
\begin{center}
\caption{GLUE task descriptions and statistics. The second and fourth column denotes the number of training examples and the number of classes. Note that STS-B is a regression task. }
\label{tbl:glue_description}
\begin{tabularx}{\linewidth}{lllp{2.5cm}LL}
\toprule
Corpus & $|\mbox{Train}|$ & $|\mbox{Label}|$ & Task & Metric(s) & Domain \\
\midrule
\multicolumn{6}{c}{Single-Sentence Classification}\\
\midrule
CoLA & 8.5k & 2 & acceptibility & Matthews corr. & misc.\\
SST-2 & 67k & 2 & sentiment & accuracy & movie reviews\\
\midrule
\multicolumn{6}{c}{Sentence Similarity/Paraphrase}\\
\midrule
MRPC & 3.7k & 2 & paraphrase & accuracy & news \\
STS-B & 5.7k & - & similarity &  Spearman corr. & misc.\\
QQP & 364k & 2 & similarity & accuracy & social QA questions \\
\midrule
\multicolumn{6}{c}{Natural Language Inference (NLI)} \\
\midrule
MNLI & 393k & 3 & NLI & (mis)matched acc. & misc. \\
QNLI & 108k & 2 & QA/NLI & accuracy & Wikipedia \\
RTE & 2.5k & 2 & NLI & accuracy & misc. \\
WNLI & 634 & 2 & coreference/NLI & accuracy & fiction books\\

\bottomrule
\end{tabularx}
\end{center}
% \vspace{-0.4cm}
\end{table*}
\begin{table}[h]
\caption{Hyperparameter search space in fine-tuning.}
\centering
\begin{tabular}{lc}
\toprule
\textbf{Hyperparameters} & \textbf{Base}  \\
\midrule
Sequence Length & 256\\
Optimizer & Adam \\
Peak Learning Rate & \{5e-5,1e-4, 3e-4\} \\
Max Epochs & \{2,3,5,10\}\\
Batch size & \{16, 32\}  \\
Learning rate decay & Linear\\
Weight Decay & \{0, 0.01\}  \\
Warm-up Proportion & \{6 \%, 10 \%\}  \\
Adam $\epsilon$ & 1e-6\\
Adam $\left(\beta_1, \beta_2\right)$ & $(0.9,0.98)$ \\
% Clip Norm & $-$ \\
Gradient Clipping & $1.0$  \\
Dropout & $0.1$  \\
\bottomrule
\end{tabular}
\label{table:fine-tune}
\end{table}

\end{document}